\documentclass[letterpaper, 10 pt, conference]{ieeeconf} 
\IEEEoverridecommandlockouts                              %
\usepackage[pdftex]{graphicx}
\usepackage{url}
\usepackage{hyperref}

\usepackage[usenames, dvipsnames]{color}
\usepackage{soul}
 
\usepackage{times} %
\usepackage{amsmath} %
\usepackage{amssymb}  %
\usepackage{gensymb}
\usepackage{bm}
\usepackage[per-mode=symbol]{siunitx}
\usepackage{pdflscape}
\usepackage{outlines}
\usepackage{multirow}
\usepackage{booktabs}

\usepackage{textcomp}
\newcommand{\midtilde}{\raisebox{0.5ex}{\texttildelow}}

\usepackage[backend=biber,      %
    style=ieee,
    bibstyle=ieee,              %
    sortcites=true,   
    mincitenames=1,
    maxcitenames=2,
    giveninits=true,            %
    backref=false
]{biblatex}
\renewcommand*{\bibfont}{\normalfont\small}
\usepackage{fancyhdr}
\newcommand{\mytitle}{\textbf{Submitted to appear in IEEE ICRA 2023.}}

\title{\LARGE \bf
Design of a Multimodal Fingertip Sensor for Dynamic Manipulation
}

\author{Andrew SaLoutos$^{1}$, Elijah Stanger-Jones$^{1}$, Menglong Guo$^{1}$, Hongmin Kim$^{1}$, and Sangbae Kim$^{1}$ %
\thanks{$^{1}$Authors are with the Biomimetic Robotics Laboratory at the Department of Mechanical Engineering, Massachusetts Institute of Technology (MIT), Cambridge, MA, 02139, USA. {\tt\small\url{saloutos@mit.edu}}}
\thanks{This work was supported by the Advanced Robotics Lab of LG Electronics Co., Ltd.} }

\begin{document}

\bstctlcite{IEEEexample:BSTcontrol}

\maketitle
\thispagestyle{fancy}
\fancyhf{}		%
\fancyfoot[L]{\normalfont \sffamily  \scriptsize \mytitle}		%
\addtolength{\footskip}{-10pt}    %
\pagestyle{empty}

\begin{abstract}
We introduce a spherical fingertip sensor for dynamic manipulation.
It is based on barometric pressure and time-of-flight proximity sensors and is low-latency, compact, and physically robust. 
The sensor uses a trained neural network to estimate the contact location and three-axis contact forces based on data from the pressure sensors, which are embedded within the sensor's sphere of polyurethane rubber.
The time-of-flight sensors face in three different outward directions, and an integrated microcontroller samples each of the individual sensors at up to 200~\si{Hz}.
To quantify the effect of system latency on dynamic manipulation performance, we develop and analyze a metric called the \textit{collision impulse ratio} and characterize the end-to-end latency of our new sensor.
We also present experimental demonstrations with the sensor, including measuring contact transitions, performing coarse mapping, maintaining a contact force with a moving object, and reacting to avoid collisions.
\end{abstract}

\section{Introduction} \label{sec:intro}

Humans can perform fast, robust, and dynamic manipulation tasks, while modern robots are often constrained to quasi-static manipulation scenarios.
While this is partly due to limitations in actuation, robot hands can be as fast and strong as human hands~\cite{grebenstein2011dlr,grebenstein2010antagonistically,kim2019fluid,shiokata2005robot,imai2004dynamic,furukawa2006dynamic,namiki2003development}. 
More importantly, human fingertips have dense concentrations of different tactile afferents to detect stimuli at broad frequency ranges, varying amplitudes, and low latencies~\cite{dahiya2009tactile}.
In contrast, modern robots often have slow or non-existent tactile sensing.
Instead, they typically rely on vision systems for planning and feedback, but cameras can become occluded during critical manipulation steps. 
In general, current tactile sensors cannot yet enable human-level manipulation performance, even when paired with nimble, dexterous fingers such as those developed by Bhatia~\cite{bhatia2019direct} and Lin~\cite{lin2021exploratory}.
\par
To achieve robust and dynamic manipulation, robots should be able to quickly sense and react to a diverse set of potential fingertip contact interactions. 
Valuable contact data can come from many directions relative to a robot's fingertips, especially in cluttered environments, and sensors should be able to capture as much of this data as possible.  
Furthermore, the sensing speed, for contact forces, contact locations, and non-anthropomorphic pre-touch proximity data, is critical for effectively controlling fingers.
Each of these data streams are necessary to avoid collisions, minimize contact forces, and control changing contact modes with objects.  
\par
\begin{figure}[t]
\centering
\includegraphics[width=\linewidth]{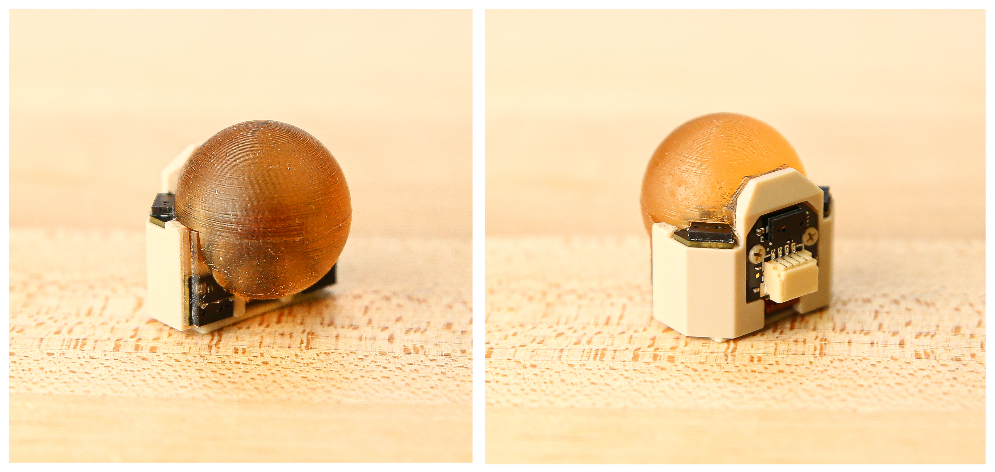}
\caption{\textbf{Multimodal fingertip sensor for dynamic manipulation.} The sensor uses barometric pressure sensors and a neural network to estimate 3-D contact forces and 2-D contact locations.
Time-of-flight proximity sensors measure distances from the three outward faces.}
\label{fig:sensors}
\vspace{-4mm}
\end{figure}

We present a compact multimodal fingertip sensor designed for dynamic manipulation tasks, building on earlier work~\cite{epstein2020bi}.
The sensor, shown in Fig.~\ref{fig:sensors}, is low-latency, physically robust, and can be easily attached to the tips of robotic fingers.
To quantify the effects of system latency on dynamic manipulation performance, we develop a metric called the \textit{collision impulse ratio}.
Finally, we present several demonstrations with our new sensor for measuring contact transitions, contact following, and coarse multimodal mapping of environments and object surfaces.

\section{Related Work} \label{sec:related}

\begin{table*}[t!]
\caption{Tactile Sensor Comparison}
\label{tab:sensor_comp}
\centering
\setlength\tabcolsep{4pt} %
\resizebox{\textwidth}{!}{
    \begin{tabular}{ l  l  l  c  c  c  c  c  c } 
     \multirow{2}{*}{\textbf{Sensor}} & \multirow{2}{*}{\textbf{Technology}} & \multirow{2}{*}{\textbf{Output}} & \textbf{Sample} &  \multirow{2}{*}{\textbf{Area}} & \textbf{Spatial} & \textbf{Force} & \textbf{Force} & \textbf{Proximity}\\
      & &  & \textbf{Rate}  & & \textbf{Resolution} & \textbf{Range} & \textbf{Resolution} & \textbf{Range}\\
     \toprule
    Human finger~\cite{dahiya2009tactile, gonzalez20152} & - & Contact force, location, texture & DC-1~\si{kHz} & - & 1~\si{mm} & 40~\si{N} & 0.06~\si{N}~$^a$ & -\\ 
     \hline
    BioTac~\cite{biotac-manual,narang2021interpreting} & Fluid-based sensing & Contact force and location & 100~\si{Hz} & 484~\si{mm}$^2$ & 2.4~\si{mm} & 10~\si{N} & 0.26~\si{N}~$^b$ & -\\ 
     \hline
    GelSight-based~\cite{yuan2017gelsight,taylor2022gelslim} & Camera with soft elastomer & Tactile RGB image & 30~\si{Hz}$^*$  & 252~\si{mm}$^2$ & 30 $\mu$m & 22~\si{N} & 1.9~\si{N}~$^b$ & -\\ 
     \hline
    Soft-bubble~\cite{kuppuswamy2020soft} & Camera with inflated elastomer & Tactile depth image & 45~\si{Hz}$^*$ & - & - & - & - & -\\ 
     \hline
    Insight~\cite{sun2022soft} & Camera with soft elastomer & 3D force map & 11~\si{Hz} & 4800~\si{mm}$^2$ & 0.4~\si{mm} & 2~\si{N} & 0.03~\si{N}~$^c$ & - \\ 
     \hline
    ReSkin~\cite{pmlr-v164-bhirangi22a} & Magnetic field & Contact force and location & 400~\si{Hz} & 400~\si{mm}$^2$ & 1~\si{mm} & 2.5~\si{N} & 0.2~\si{N}~$^a$ & -\\ 
     \hline
    Lancaster et al~\cite{lancaster2019improved} & Time-of-flight & Contact force and proximity & 30~\si{Hz} & \midtilde 600~\si{mm}$^2$ & - & 10~\si{N} & - &  0-50~\si{mm}\\ %
     \hline
    Ours (previous)~\cite{epstein2020bi} & Pressure & Contact force and location & 200 Hz & 184~\si{mm}$^2$ & \midtilde 1.11~\si{mm} & 25~\si{N} & 0.65~\si{N}~$^b$ & -\\ %
     \hline
    \textbf{Ours (this paper)} & Pressure, time-of-flight & Contact force, location, and proximity & 200 Hz & 406~\si{mm}$^2$ & \midtilde 1.31~\si{mm} & 25~\si{N} & 1.58~\si{N}~$^b$ & 0-150~\si{mm}\\ %
     \bottomrule
     \\ \multicolumn{9}{l}{$^*$ Sample rate is based only on camera frame rate, additional latency was not reported.}
     \\ \multicolumn{9}{l}{$^{a,b,c}$ Reported force resolution as $^a$ minimum distinguishable force or $^b$ root-mean-squared-error (RMSE) or $^c$ median error relative to an evaluation dataset.} \\
    \end{tabular}
}
\vspace{-6mm}
\end{table*}

In this section, we review relevant tactile sensors, address applications of proximity sensing to tactile sensing, and discuss our previous sensor design.
Table~\ref{tab:sensor_comp} compares tactile sensors using various sensing technologies and modalities, including our new sensor. 
More comprehensive reviews of tactile sensing based on different technologies, including proximity sensing, can be found in~\cite{dahiya2009tactile, kappassov2015tactilereview, shimonomura2019tactile, navarro2021proximity}.

\subsection{Tactile Sensing}
Tactile sensors can be grouped two categories: vision-based and non-vision-based.
While traditional force/torque sensors, such as those from ATI~\cite{ati-site}, can record high-bandwidth force and torque data, they are not well-suited to tactile sensing because they are bulky, heavy, and prone to resonance and inertial noise.
Non-vision-based tactile sensors have been developed using resistive~\cite{van_den_2009, fernandez2014vibration, choi2022tactile}, capacitive~\cite{da_rocha_2009, gruebele2020stretchable, huh2020dynamically}, barometric pressure~\cite{tenzer_jentoft_howe_2014, piacenza2018data, epstein2020bi}, and magnetic field~\cite{pmlr-v164-bhirangi22a} modalities, among others. 
Sensors with resistive and capacitive measurements are cheap and robust.
They can cover large sensing areas, but they are also noisy and low resolution, which limits their effectiveness for fingertip sensors. 
The commercially-available BioTac fingertip sensor~\cite{biotac-manual} measures fluid pressure to mimic the modalities of the human fingertip, but it is expensive and fragile.
The ReSkin sensor, which uses magnetometers, has a very high sample rate but a low maximum normal force and is limited to a 2-D planar profile~\cite{pmlr-v164-bhirangi22a}.
\par
Vision-based tactile sensors are constructed around off-the-shelf cameras and reflective elastomer skins~\cite{shimonomura2019tactile} and operate by capturing images of imprints of touched objects. 
The GelSight sensor~\cite{yuan2017gelsight} is one of the most common vision-based sensors; 
its design has been modified for slimmer fingers~\cite{taylor2022gelslim, ma2019dense}, integration into tips of commercial robot hands~\cite{lambeta2020digit}, 3-D sensing of surfaces~\cite{romero2020soft, padmanabha2020omnitact, sun2022soft}, and pre-contact sensing with semi-transparent skin surfaces~\cite{yamaguchi2016combining, hogan2022finger, yin2022multimodal}.
Other vision-based sensors use depth cameras instead of RGB cameras~\cite{alspach2019soft, kuppuswamy2020soft} to obtain tactile point clouds. 
\par
Vision-based tactile sensors have been used for a wide range of manipulation tasks, from cable manipulation~\cite{she2021cable} to iterative re-grasping~\cite{calandra2018more}.
They have excellent spatial resolution and are well-suited for powerful learning approaches for estimation and control~\cite{yuan2017gelsight, wang2020swingbot, kuppuswamy2020soft}.
However, they suffer from severe limitations during dynamic applications, as the amount of collected image data introduces significant latency in transmission and processing.
First, while the frame rates for the cameras used in these sensors can reach 90 fps, additional latencies of 40~\si{ms} to 70~\si{ms} are reported in~\cite{romero2020soft, wilson2020design}, which can drop the bandwidth to 20 Hz or lower. 
Second, control algorithms capable of dealing with large amounts of data, such as convolutional neural networks, are often very intensive to evaluate, especially when using multiple sensors.
As a result, while the sensors can be trained to estimate contact forces, applications are generally restricted to slow, quasistatic, and low-force interactions.

\subsection{Proximity Sensing}
Proximity sensing is complementary to typical sensing modalities of vision and touch and has commonly been used in robotics applications for safe human-robot interaction and reactive grasping~\cite{navarro2021proximity}. 
Typically, capacitive or optical sensing technologies are used, but optical sensors can provide longer sensing ranges. 
Optical proximity sensors have been used in robotic manipulation for finger shaping control~\cite{koyama2013pre, koyama2016integrated, koyama2018high} and coarse mapping of object surfaces~\cite{patel2016integrated, yang2017pre, kang2017capacitive}. 
\par
Optical proximity sensors that measure the intensity of reflected light~\cite{hsiao2009reactive, hasegawa2010development, koyama2013pre, konstantinova2015force, koyama2016integrated, patel2016integrated, yamaguchi2018gripper, koyama2018high} are easy to deploy and have fast sample rates.
However, their outputs are highly dependent on object properties, which makes them difficult to calibrate.
Optical time-of-flight sensors have slower sample times, but they return precise measurements over broader distances and are not as dependent on object properties.
Miniature versions of these sensors, specifically the VL6180X from STMicroelectronics, have recently been integrated into many systems~\cite{yang2017pre, huang2018visionless, sasaki2018robotic, lancaster2019improved, hasegawa2020online}. 
Multimodal sensors have been created using optical proximity sensing~\cite{patel2016integrated, lancaster2019improved}, but the outputs are still strongly dependent on object properties.

\subsection{Previous Sensor Design}
This work improves on our previous sensor~\cite{epstein2020bi}, which used barometric pressure sensors embedded in polyurethane rubber to estimate contact force and location. 
The sensing area of the previous design is limited due to its hemispherical shape.
While the sensor performs well for antipodal grasps on a 1-DoF gripper~\cite{saloutos2022fast}, the small sensing area limits the effectiveness of more nimble fingers.
\par 
Our new version uses a spherical design to expand the sensing area and allow contact sensing over a $180\degree$ range without adding more pressure sensors or decreasing the sampling rate.
We also include time-of-flight proximity sensors on the edges of the sensor base to measure pre-contact distances.
By placing proximity sensors outside the rubber and using pressure sensors inside the rubber, we avoid compromising the proximity modality, unlike other multimodal sensors~\cite{patel2016integrated,lancaster2019improved,segil2019multi}.

\section{Importance of Sensing Bandwidth} \label{sec:bw}

Unintended collisions can ruin carefully constructed manipulation plans by causing an object to escape the fingers' grasp, especially if the system has too much latency to prepare for and react to the contact appropriately.
So, low system latency, through actuation bandwidth, sensing bandwidth, and fast information processing, is critical for fast, reactive, and robust manipulation performance. 
However, the effect of latency, especially during collisions and contacts, is not captured in traditional metrics for manipulation or dynamic capabilities. 
For manipulators, metrics such as manipulability or dynamic manipulability do not consider behavior during collisions~\cite{yoshikawa1985manipulability, yoshikawa1985dynamic}.
The impact mitigation factor, or IMF, considers the passive dynamics of floating-base systems during collisions but does not address system latency~\cite{Wensing2017ko}.
Bhatia~\cite{Bhatia-2022-131694} creates a collision reflex metric to characterize the transparency of closed-loop robotic systems based on the total impulse during a collision between the robot and the environment.
However, their collision model includes non-physical velocity jumps and does not explicitly include latency.
Here, we model the full collision dynamics and explicitly include the system latency to show that it is critical to overall system performance and impulse minimization.

\begin{figure}[t]
\centering
\includegraphics[width=\linewidth]{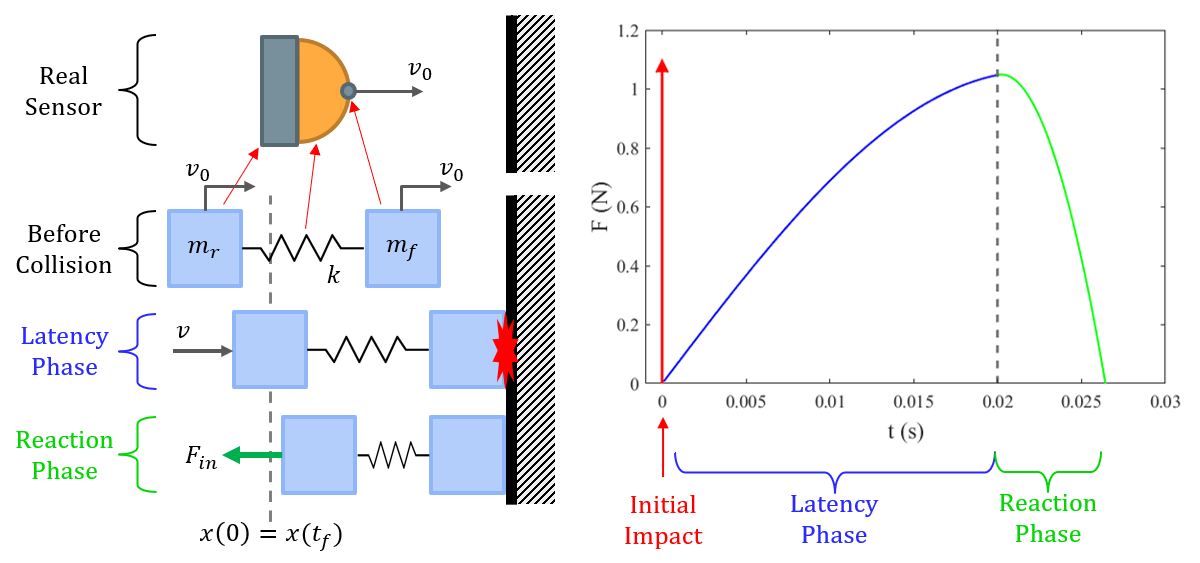}
\caption{ \textbf{Collision model diagram.} \textit{Left:} The sensor collision model consists of two masses, connected by a spring, impacting a rigid object with an initial velocity. After a delay due to latency, the control system can react to the collision. \textit{Right:} Three distinct regions of the force plot correspond to three phases of the collision.}
\label{fig:collision}
\vspace{-4mm}
\end{figure}

\begin{figure}[t]
\centering
 \includegraphics[width=\linewidth]{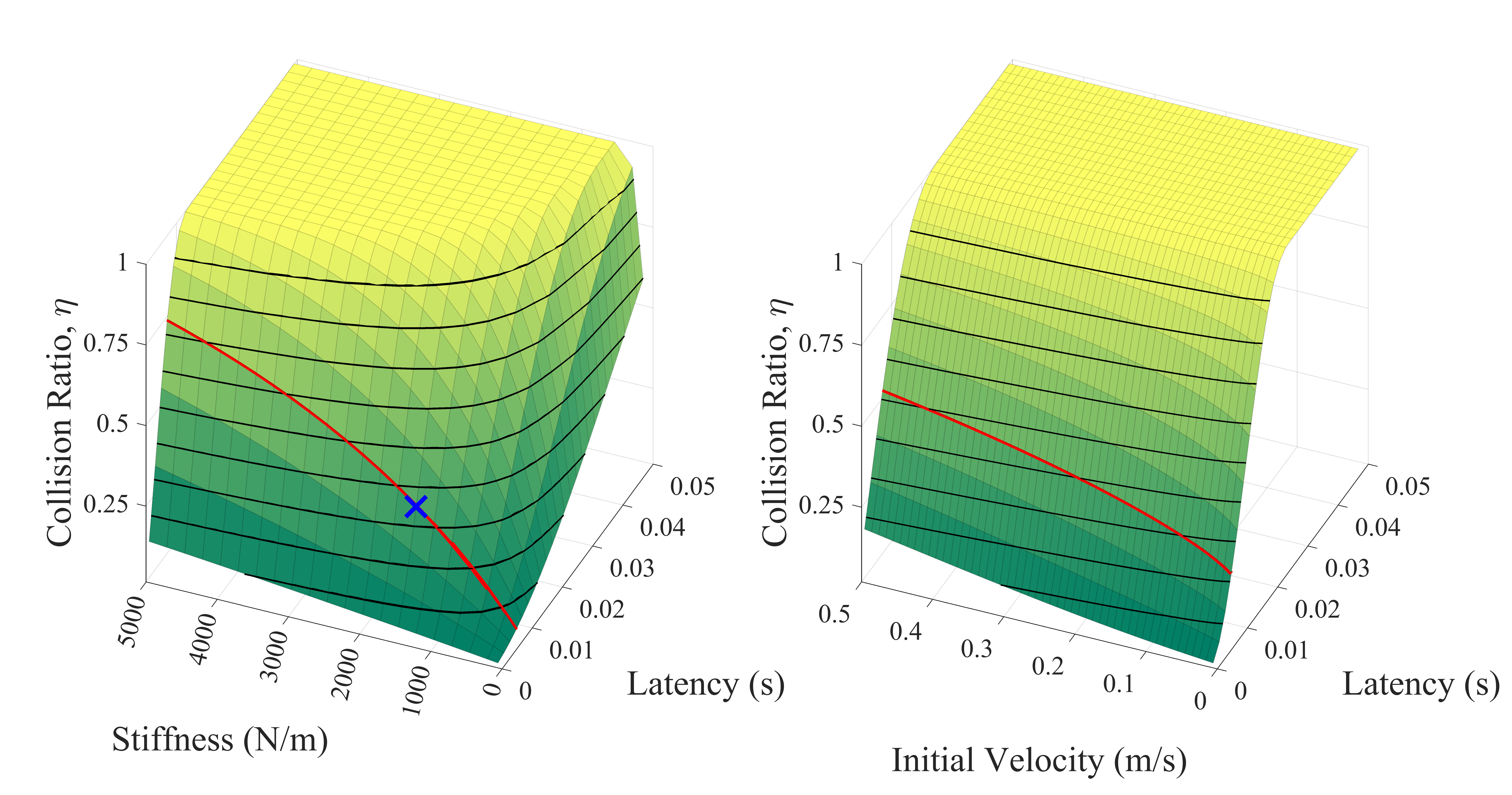}
\caption{ \textbf{Collision impulse ratios.} \textit{Left:} Collision impulse ratio, $\eta$, for varied stiffnesses and latencies. For high stiffnesses, reducing latency will reduce the ratio. \textit{Right:} $\eta$ for varied initial velocities and latencies. Initial velocity does not have a large effect on the ratio compared to the latency. For both plots, red slice corresponds to the actual sensor latency of 7~\si{ms}, as reported in Section~\ref{sec:latency}. The blue `x' marks the measured stiffness of the sensor. Nominal simulation parameters: $[m_f, m_r, t_l, k, v_0, F_{in}] = [0.005\si{kg}, 0.1\si{kg}, 7\si{ms}, 1500\frac{\si{N}}{\si{m}}, 0.15\frac{\si{m}}{\si{s}},10\si{N}]$.}
\label{fig:impulse}
\vspace{-6mm}
\end{figure}

\par
Our finger collision model is shown in Fig.~\ref{fig:collision}.
It is based on the model in~\cite{Bhatia-2022-131694}. 
The model consists of a fingertip mass, $m_f$, and a finger mass, $m_r$, connected by a spring with stiffness $k$, which collide with a rigid object. 
In our model, the fingertip mass is nearly zero, representing the sensor's surface, and the finger mass is the effective mass of the end of an individual robot finger. 
The stiffness represents the lumped stiffness of the finger in series with the rubber of the sensor. 
Before the collision, both masses have an initial velocity $v_0$. 
\par
The collision starts with a plastic impact between the fingertip mass and the rigid object, at which point the displacement of the finger mass, $x(t)$, is set to $x(0) = 0$. 
This initial impact generates an impulse that, when detected, indicates to the control system that a collision has occurred.
The collision ends at $t_f$ when the displacement of the finger mass returns to $0$, such that $x(t_f) = 0$. 
The system latency, $t_l$, defines a delay from the start of the collision, after which the finger actuator can react to the initial impulse, exerting a force $F_{in}$ on the finger mass. 
During the delay period, we assume the mass-spring system behaves according to its unforced dynamics.
A fundamental assumption in our model is that the initial impulse is detectable after the latency period, regardless of incoming velocity.
This assumption depends strongly on the sensitivity of the force sensor; realistically, the magnitude of the initial impulse needs to be above the product of the sensor's minimum force resolution and sample time for it to be detected.
\par
Based on our collision model, the displacement of the finger mass during a collision is given by
\begin{equation}
    x(t) = 
    \begin{cases}
        \frac{v_0}{\omega_0}\sin(\omega_0t), & 0 \leq t < t_l \\
        (x_l - \frac{F_{in}}{k})\cos(\omega_0(\tilde{t})) \\ 
        \quad + \frac{v_l}{\omega_0}\sin(\omega_0\tilde{t}) + \frac{F_{in}}{k}, & t_l \leq t \leq t_f
    \end{cases}
\end{equation}
\noindent where $\tilde{t} = t - t_l$, $x_l = \frac{v_0}{\omega_0}\sin(\omega_0 t_l)$, and $v_l = v_0\cos(\omega_0 t_l)$.
\par
After the initial fingertip collision, the system's natural frequency is given by $\omega_0 = \sqrt{\frac{k}{m_r}}$ and the total force on the finger mass is given by $F(t) = -kx(t)$.
The total impulse, obtained by integrating the force during the collision, is
\begin{equation} \label{eq:impulse}
\begin{aligned}
    I &= m_f v_0 + m_r v_0(1-\cos(\omega_0 t_f)) \\
                    &\quad + F_{in}(t_f - t_l - \frac{\sin(\omega_0(t_f-t_l))}{\omega_0}.
\end{aligned}
\end{equation}
\par
To isolate the effect of varied system latencies on the total impulse, we normalize by the total impulse for the system as if there were no control input, i.e. $F_{in} = 0$ in~\eqref{eq:impulse}; essentially, this is normalizing by the natural dynamics. 
This ``natural'' impulse is given by
\begin{equation} \label{eq:impulse_natural}
    I_{N} = m_fv_0 + m_rv_0(1-\cos(\omega_0t_f)) 
\end{equation}
We call our normalized total impulse the \textit{collision impulse ratio}, represented by $\eta$:
\begin{equation}
    \eta = \frac{I}{I_{N}}
\end{equation}
This ratio represents how much of the collision impulse a system can avoid due to its control effort after the latency period.
It can vary from 0 to 1, with smaller ratios being more desirable.
\par
Fig.~\ref{fig:impulse} shows plots of the collision impulse ratio, $\eta$, for various system latencies, stiffnesses, and initial velocities. 
The isolines of $\eta$ are shown in bold black to give a better sense of the shape of the surfaces. 
The left plot shows that low stiffness and low latency can each reduce the collision impulse ratio.
However, high stiffness is desirable for the fast, high-bandwidth control actions necessary in dynamic manipulation scenarios.
So, reducing latency is the most practical method for minimizing impulses during collisions.
The right plot shows that the collision impulse ratio is nearly invariant to initial velocities, due to the normalization by the natural impulse.
However, the ratio is sensitve to the latency for any initial velocity, even ones close to zero.
From this analysis, we can conclude that minimizing the system latency, independent of other parameters, is crucial for good manipulation performance. %
\par
In real-world systems, the total latency used in this analysis can come from actuation, sensing, and data processing.
While high actuation bandwidth is essential for fast and dynamic manipulation, fully utilizing fast actuators also requires fast sensors. 
In particular, minimizing the latency is useful for tactile sensors for quickly detecting and reacting to collisions.
In the case of our sensor, the contact force measurement is obtained at the maximum sample rate of the individual pressure sensors, and the estimation neural network is small and fast to evaluate.
Using pre-touch data from the proximity sensors effectively reduces our sensor latency further by allowing fingers to prepare for collisions before they occur, as demonstrated in Section~\ref{sec:react}.

\section{Sensor Design} \label{sec:design}

\begin{figure}[t]
\centering
\includegraphics[width=\linewidth]{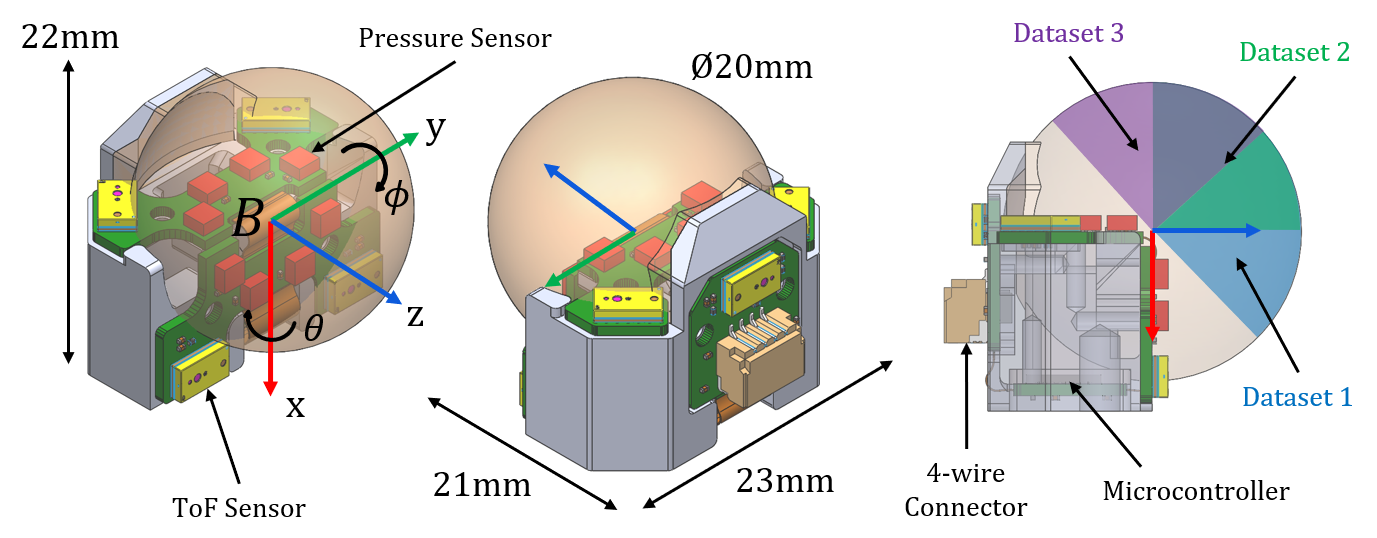}
\caption{ \textbf{Sensor design.} The time-of-flight sensors are shown in yellow and the pressure sensors are shown in red. The two contact angles, $\theta$ and $\phi$, correspond to positive rotations about the x- and y-axes of the sensor base frame, $B$. The portions of the sensor surface covered by each training dataset are shown in blue, green, and purple.}
\label{fig:sensor_cad}
\vspace{-4mm}
\end{figure}

Our sensor design is compact, physically robust, and features high-bandwidth contact force, location, and proximity sensing modalities.
Fig.~\ref{fig:sensor_cad} shows detailed views of the new sensor. 
The complementary proximity and contact sensing modalities allow us to have accurate distance measurements without compromising our contact sensing bandwidth.
The expanded contact area allows for a broad and continuous sensing range for diverse motions, including antipodal grasping, forward prodding, and finger extension. 
The time-of-flight proximity sensing along different axes allows for detecting and tracking objects between and in front of the fingertips.
\par
The sensor consists of a rigid-flex PCB, a rigid base piece, and a cast rubber dome. 
It can sense contacts over a $180\degree$ by $90\degree$ region of the dome's surface, corresponding to $-45\degree$ to $+135\degree$ about the base frame's y-axis and $\pm45\degree$ about the base frame's x-axis, and can measure shear forces of $\pm15$~\si{N} and normal forces up to $25$~\si{N}.
The time-of-flight proximity sensors have a range of $10$~\si{mm} to approximately $150$~\si{mm}.
\par
The rubber dome is 20~\si{mm} in diameter and has a usable sensing area of over 400~\si{mm}$^2$.
Within the cast rubber are eight BMP384 barometric pressure sensors, which can be sampled at 200 Hz. 
These sensors are chosen for their small footprint, just 2~\si{mm} by 2~\si{mm}, and fast sampling rate.
The pressure sensors are evenly split between the first two boards of the rigid-flex PCB, which are mounted facing perpendicular directions. 
Distributing the individual sensors across two boards enables us to expand the sensing area compared to~\cite{epstein2020bi}. 
\par
Five VL6180X time-of-flight proximity sensors are placed around the rubber dome and on the back sensor surface to measure distances along three outward directions from the fingertip.
The sample rate of time-of-flight sensors is not deterministic, but the typical update rate is roughly 100 Hz. 
These sensors are chosen because they are compact, fast, and have a long sensing range.
\par
An STM32F334 microcontroller and a CAN transceiver are included on the sensor PCB to sample the individual sensors and send data back to the control system. 
A cable consisting of four flexible and thin 30AWG wires powers and communicates with the sensor, making it easier to route wires from the tip of the finger.
The rigid base piece is machined from PEEK plastic, with two threaded inserts for mounting to robot fingers or other hardware. 
The rubber dome is cast from Smooth-On VytaFlex 20 polyurethane rubber.
The entire sensor is approximately 23~\si{mm} x 22~\si{mm} x 24~\si{mm} and has a total mass of about 10~\si{g}.

\section{Data Collection and Training Results} \label{sec:train}

\begin{figure*}[t]
\centering
\includegraphics[width=\linewidth]{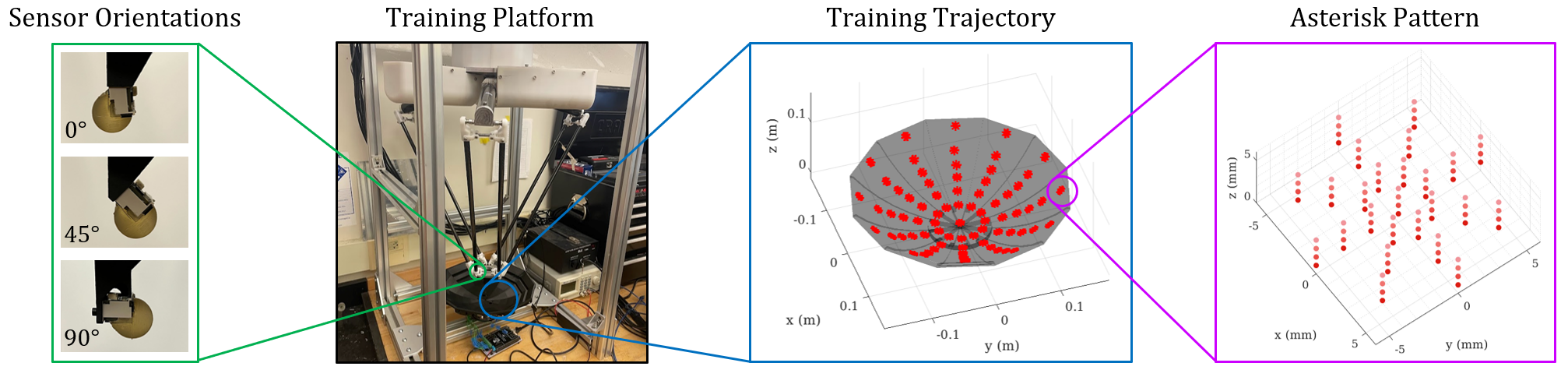}
\caption{\textbf{Sensor training setup.} The training platform consists of the delta robot, an ATI sensor, and the contact bowl. Data is collected with the sensor in three orientations. For each inidividual dataset, the robot presses the sensor into each contact patch and executes a multi-layered asterisk pattern, exerting the full range of shear and normal forces.}
\label{fig:train_setup}
\vspace{-4mm}
\end{figure*}

A neural network is trained on the data from the eight internal pressure sensors to estimate the 2-D contact location on the spherical surface and the corresponding 3-D contact force.
The network has eight inputs and three fully connected layers of 12, 64, and 64 nodes with ReLu activations. 
There are five outputs, so the model can be represented as
\begin{equation} \label{eq:nn}
    [F_x, F_y, F_z, \theta, \phi]^T = f(s_1, s_2, s_3, s_4, s_5, s_6, s_7, s_8)
\end{equation}
where $s_i$ are the individual pressure readings.
\par
The contact angles $\theta$ and $\phi$ define the contact location on the sphere surface relative to the sensor base frame, shown in Fig.~\ref{fig:sensor_cad}.
The first contact angle, $\theta$, defines an initial rotation about the x-axis of the sensor base frame. 
The second contact angle, $\phi$, defines a subsequent rotation about the y-axis of the sensor base frame. 
The SE(3) transform between the sensor base frame and the contact frame on the sensor surface is given by
\begin{equation} \label{eq:contact_transform}
\begin{split}
\boldsymbol{T}_{base,contact} &= 
\begin{bmatrix}
\boldsymbol{R}_y(\phi) & \boldsymbol{0} \\
\boldsymbol{0}^T & 1 \\
\end{bmatrix}
\begin{bmatrix}
\boldsymbol{R}_x(\theta) & \boldsymbol{0} \\
\boldsymbol{0}^T & 1 \\
\end{bmatrix}
\begin{bmatrix}
\boldsymbol{I}_3 & \boldsymbol{p}_0 \\
\boldsymbol{0}^T & 1 \\
\end{bmatrix}
\\
\boldsymbol{p}_0 &= \begin{bmatrix} 0 & 0 & r_{sensor} \end{bmatrix}^T
\end{split}
\end{equation}
where $r_{sensor} = 10~\si{mm}$ is the sensor radius. 
The three-axis force vector, $F_x$, $F_y$, and $F_z$, is defined in the contact frame. 
The $z$ component of the vector is the normal force, and the $x$ and $y$ components are the shear forces.
The z-axis of the contact frame is the contact normal vector. 
\par
The sensor is fixed to the end-effector of a delta robot and repeatedly brought into contact with a rigid bowl attached to an ATI Delta SI-660-60 sensor to collect training data for the neural network estimator.
The end-effector can move in $x$, $y$, and $z$ directions, and the bowl has planar patches corresponding to pairs of sensor contact angles.
During each period of contact between the sensor and the bowl, the robot moves in a layered asterisk pattern to induce different levels of shear and normal forces at each contact location. 
Due to the high coefficient of friction between the sensor rubber and the bowl, there is little slip during the contact periods. 
If slip does occur, however, the sensor contact angles do not change because each patch of the bowl is planar.
Fig.~\ref{fig:train_setup} shows the data collection setup with a detailed view of the contact bowl and an example of the asterisk trajectory for each contact patch. 
\par
To cover the expanded sensing surface of the new sensor, we collect three separate datasets and combine them to form the training dataset.
Each individual dataset corresponds to a $90\degree$ cone of the sensor surface, so the sensor needs to be rotated by $0\degree$, $-45\degree$, and $-90\degree$ about the y-axis of the sensor base frame to cover the entire sensing surface. 
The regions corresponding to each of the three individual datasets are illustrated in Fig.~\ref{fig:sensor_cad}, and Fig.~\ref{fig:train_setup} shows each of the three sensor training orientations.
During training, the applied shear forces range from approximately $-15$~\si{N} to $+15$~\si{N}, the normal force ranges from approximately $0$~\si{N} to $-25$~\si{N}, the contact angle $\theta$ ranges from $-45\degree$ to $+45\degree$, and the contact angle $\phi$ ranges from $-135\degree$ to $+45\degree$.
\par
The full dataset consists of roughly 350,000 data points. 
We randomly select 90\% of the dataset for training, and the remaining 10\% of the dataset is used for testing. 
The network is trained using Keras with the Adam optimizer for ten epochs with a batch size of ten.
Collecting the entire training dataset for one sensor takes roughly 20 minutes of operator setup time and 3 hours of robot run time.
Training the estimator network takes less than three minutes using a desktop with an i9-11900K processor.
\par
Table~\ref{tab:training} reports the RMSE values for the training and testing datasets for two sensors manufactured according to the presented design.
Compared to the previous sensor version, the RMSE values are slightly higher, and the values for sensor \#1 are worse than those for sensor \#2. 
However, in practice we have not observed any negative effects due to the higher errors.
Furthermore, we have not noticed significant drift in the overall force and contact outputs after over 40 hours of experimental use of the sensors.
\par
A drawback of this training setup is that the contact angle parameterization and the contact bowl design require a spherical sensor surface. 
While the well-defined contact kinematics enable a smooth transfer from the training setup to real-world applications, the spherical shape is not as practical for grasping as a flatter, more oblong shape, since it results in point contacts rather than larger contact patches.

\begin{table}[t]
\caption{Sensor Training Results, RMSE}
\label{tab:training}
\centering
\begin{tabular}{ c  c  c  c  c } 
 \multirow{2}{*}{\textbf{Sensor}} & \multicolumn{2}{c }{\textbf{Train}} & \multicolumn{2}{c }{\textbf{Test}}  \\
 & \textit{Forces} & \textit{Angles} & \textit{Forces} & \textit{Angles} \\
 \toprule
\textbf{1} & 1.71~\si{N} & 0.153rad & 1.72~\si{N} & 0.154rad \\ %
 \midrule
\textbf{2} & 1.42~\si{N} & 0.108rad & 1.43~\si{N} & 0.109rad \\ %
\bottomrule
\end{tabular}
\vspace{-4mm} 
\end{table}

\section{Experimental Demonstrations} \label{sec:demo}

\subsection{Sensor Latency and Contact Transitions} \label{sec:latency}
To characterize the end-to-end latency of the contact force sensing, we mounted the sensor to the training platform and repeatedly pressed it into the surface of the ATI sensor, collecting data at 1~\si{kHz} across the range of force capabilities.
Similarly, to measure the latency of the proximity sensor, we mounted the sensor to an instrumented stage and moved it back and forth repeatedly, collecting data at 2~\si{kHz}.
We estimated the end-to-end latency for both sensing modalities by calculating the time-domain shift that maximized the cross-correlation between the ground truth and processed sensor signals.
This measured latency includes all communication, processing and mechanical delays.
The calculated contact force latency is approximately 7~\si{ms} averaged across the usable force ranges, and the proximity latency is approximately 4~\si{ms}.
\par
\begin{figure}[t]
\centering
\includegraphics[width=\linewidth]{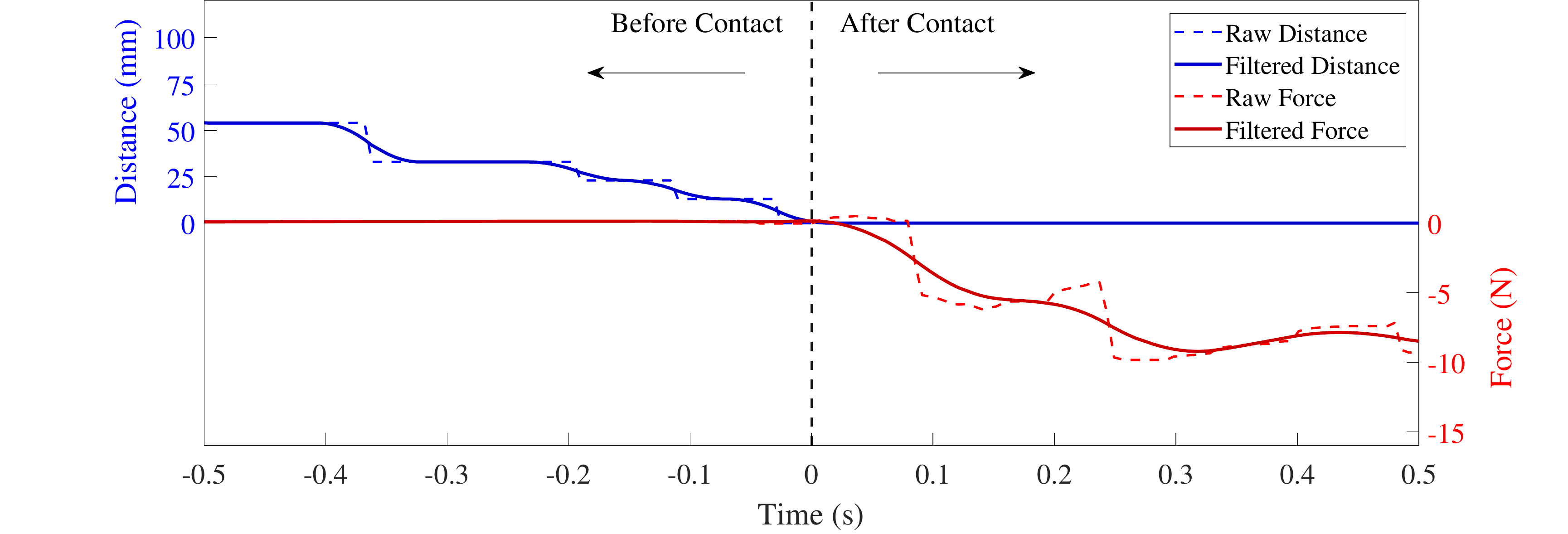}
\caption{\textbf{Example contact transition.} Before the sensor comes into contact with a block, the proximity measurement, shown in blue, go to zero. After the contact, the normal force measurement, shown in red, becomes non-zero. The dashed lines in the figure correspond to the raw data. The solid lines correspond to filtered data. Moving average windows of 7 and 15 for the proximity and force data, respectively, were applied using the \texttt{filtfilt} function in Matlab.}
\label{fig:transition}
\end{figure}

Fig.~\ref{fig:transition} shows measurements from a single proximity sensor and the sensor's estimated normal force, recorded at approximately 200~\si{Hz}, as the fingertip makes contact with a rigid block.
The lower bound of the proximity sensing range is reached as the block makes contact with the sensor surface.
Once contact is made, the contact force and location can be measured. 
There is a smooth transition between the two sensing modalities at the moment of contact.

\begin{figure}[t]
\centering
\includegraphics[width=\linewidth]{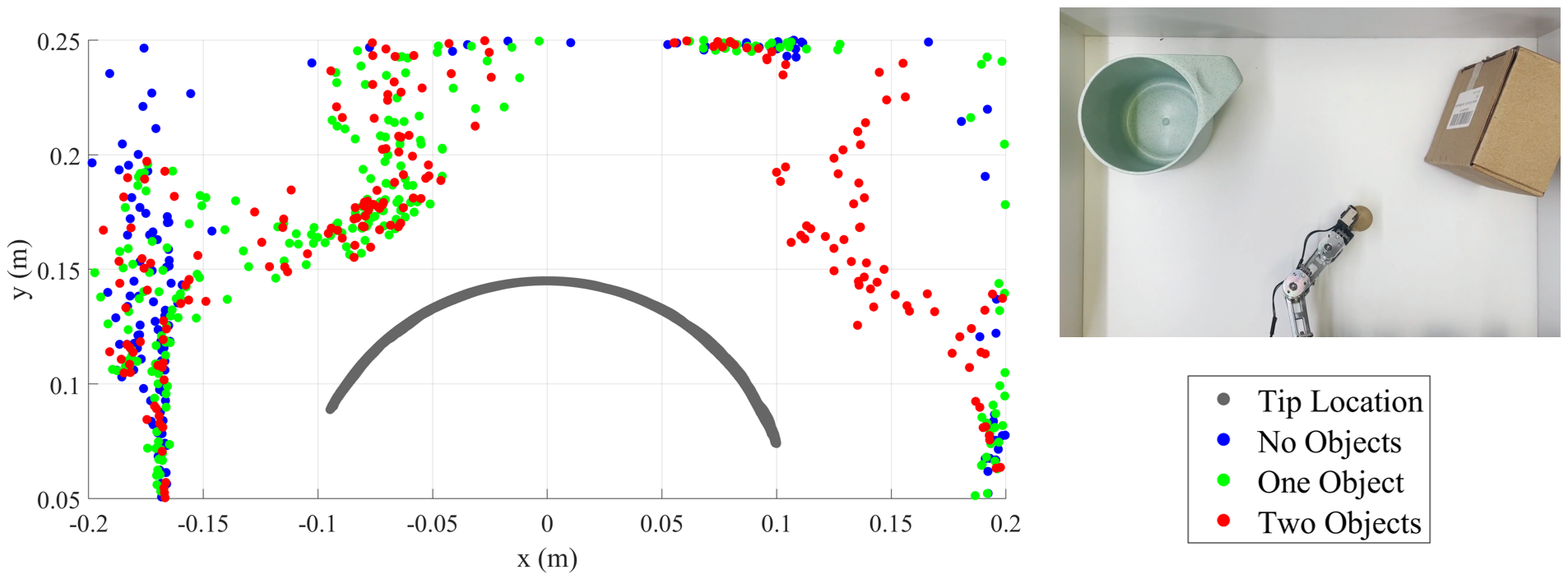} \\
\vspace{2mm}
\includegraphics[width=\linewidth]{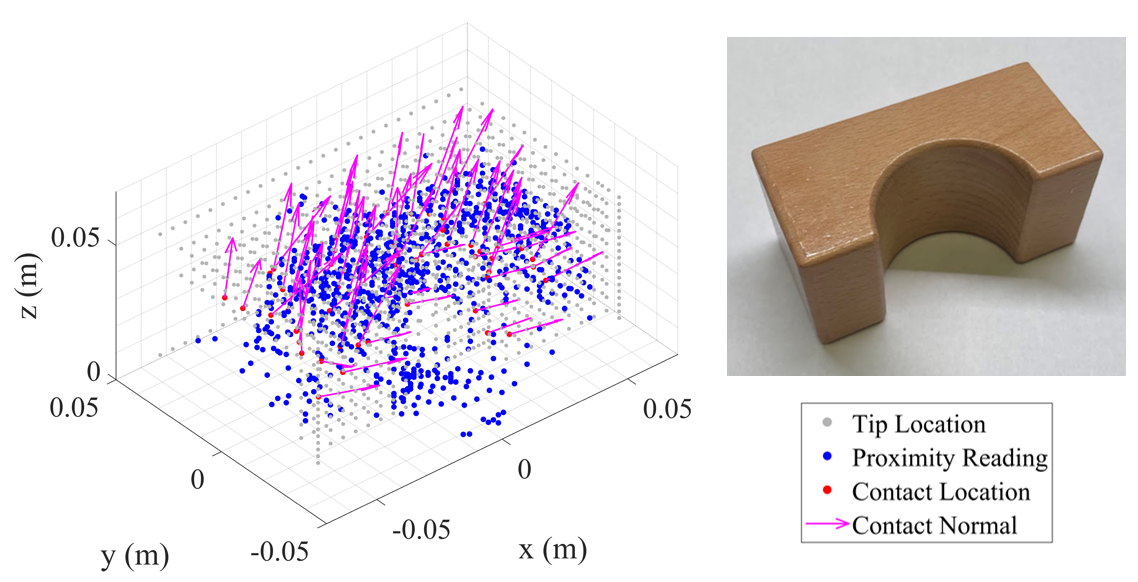}
\caption{ \textbf{Mapping experiments.} The passively collected proximity data can be used to create coarse maps of the environment or of object surfaces. For object surfaces, any contact locations and associated normal vectors can be included in the map with the proximity data.}
\label{fig:map_tap}
\vspace{-4mm} 
\end{figure}

\subsection{Mapping of Objects and the Environment}
We can use the proximity data from when the finger is in free motion to generate coarse maps of the environment near the fingertip.
For example, the upper part of Fig.~\ref{fig:map_tap} shows the actual environment, which has three walls and two objects, and a coarse map generated by sweeping the fingertip back and forth.
Across three experiment trials, we sequentially removed the objects from the environment. 
These changes are accurately reflected in the maps from the fingertip sensor for each trial.
\par
The surfaces of individual objects can also be mapped using proximity data.
If the sensor contacts the object, the map can then be updated to include the world contact location and the contact normal vector.
The lower part of Fig.~\ref{fig:map_tap} shows the combined mapping data as the fingertip was moved in 3-D space around a wooden block, intermittently making contact with the object.
Dense mapping with only contact data is sample-inefficient, but by incorporating the proximity data from between taps on the object, we can quickly create an accurate map. 
Furthermore, the proximity data could be used to reduce uncertainty in collision timing before the finger taps to prevent excessive disturbances.

\subsection{Dynamic Reactions}  \label{sec:react}
The low sensor latency enables dynamic reaction behaviors, such as stable contact following and potential fields for collision avoidance.
Real-time videos from the contact following and collision avoidance trials can be found \href{https://youtu.be/tr5UQTEbIuk}{here}\footnote{Videos can be found at https://youtu.be/tr5UQTEbIuk}.
\par
The combination of fast contact location and contact force sensing allows us to create a contact-following behavior with our sensor.
In this behavior, a user initiates contact with the fingertip sensor.
Then, feed-forward joint torques exert a constant force at the fingertip along the sensed contact normal. 
As the user maneuvers the contact surface, and thus the contact point and contact normal, the fingertip ``sticks'' stably to the contact, following it through the workspace.
Qualitatively, the fingertip behaves like a virtual magnet, using only feed-forward force control at the actuators.
\par
Proximity data can be used to create virtual potential fields around each fingertip.
Any measured distance below a specified threshold creates a spring force on the fingertip, which is tracked using feed-forward joint torque commands. 
The potential fields allow the fingers to quickly react to seen objects and avoid collisions or unexpected contacts without a long planning horizon or vision sensing.
\par
We use a combination of similar reactive behaviors, called reflexes, to achieve robust and autonomous grasping \cite{saloutos2022reflex}. 
The reflexes take advantage of our manipulation platform's low-latency sensors and fast actuation bandwidth to react to objects in the environment, compensate for inaccurate planning based on coarse vision information, and control contact interactions with grasped objects. 

\section{Conclusion} \label{sec:conc}

We have presented a new sensor that combines proximity and pressure sensing modalities to provide data before, during, and after contacts.
It is compact for easy integration into the fingertips of robotic hands, and it is well-suited to dynamic manipulation scenarios due to its low latency.
\par
We plan to deploy this fingertip sensor in our manipulation system with the final goal of achieving human-level manipulation skills.
To do this, we will need to improve the sensor sampling rate and force resolution, especially if we want to operate in the low-force regimes in which humans excel.
Additionally, vision-based planning and vision-based tactile sensing, used before first contact is made and after a grasp has been completed, respectively, are complementary to our sensor's capabilities and could help to further reduce the gap to human manipulation.  
\par
Future iterations of the sensor will include several engineering improvements. 
As commercially available sensors improve, we will upgrade to sensors with higher sampling rates to further increase our bandwidth. 
We can also improve our sensing accuracy with better casting processes for the rubber dome or different locations of the individual pressure sensors.
Finally, each sensor currently requires a uniquely trained neural net model. 
An important direction for future work on improving the usability of the sensors is to train a model that generalizes across all of the sensors that share the same shape and topology.

\printbibliography

\end{document}